\documentclass[journal,twoside]{IEEEtran}
\usepackage{setspace}
\usepackage{booktabs}
\usepackage[dvips]{epsfig}    
\usepackage{enumerate}
\usepackage{graphicx}         
\usepackage{remreset}
\usepackage{subfigure}
\usepackage{float}
\usepackage{hyperref}
\makeatletter \@removefromreset{figure}{section} \makeatother
\usepackage{enumerate, color}
\usepackage{amssymb}
\usepackage{amsmath}
\usepackage{framed}
\usepackage{mathrsfs}
\usepackage{amsmath,bm}
\usepackage{amsthm}
\usepackage{multirow}
\usepackage{array}
\usepackage{verbatim}
\usepackage{color}
\usepackage{subfigure}
\usepackage{multicol}
\usepackage[section]{placeins}
\usepackage{footnote}
\usepackage{balance}
\usepackage[keeplastbox]{flushend}
\usepackage{url}
\usepackage{algorithm}
\usepackage{algorithmic}
\usepackage{mathrsfs}
\usepackage{textcomp}

\begin{document}

\title{L1-Norm Batch Normalization for Efficient Training of Deep Neural Networks}

\author{
  Shuang Wu$^{*}$,
  Guoqi Li$^{*}$,
  Lei Deng,
  Liu Liu,
  Yuan Xie,
  and Luping Shi$^{\dag}$


\thanks{
S. Wu and G. Li contribute equally to this work. S. Wu, G. Li and L. Shi are with the Department of Precision Instrument, Center for Brain Inspired Computing Research, Tsinghua University, Beijing, 100084, China (e-mail: wus15@mails.tsinghua.edu.cn; liguoqi@mail.tsinghua.edu.cn; lpshi@mail.tsinghua.edu.cn)}
\thanks{
L. Deng, L. Liu and Y. Xie are with Department of Electrical and Computer Engineering, University of California, Santa Barbara, CA93106, USA (e-mail: leideng@ucsb.edu; liu\_liu@ucsb.edu; yuanxie@ucsb.edu)}
\thanks{
$^{\dag}$ Corresponding to: lpshi@mail.tsinghua.edu.cn}

}


\maketitle

\begin{abstract}

Batch Normalization (BN) has been proven to be quite effective at accelerating and improving the training of deep neural networks (DNNs).
However, BN brings additional computation, consumes more memory and generally slows down the training process by a large margin, which aggravates the training effort.
Furthermore, the nonlinear square and root operations in BN also impede the low bit-width quantization techniques, which draws much attention in deep learning hardware community.
In this work, we propose an L1-norm BN (L1BN) with only linear operations in both the forward and the backward propagations during training.
L1BN is shown to be approximately equivalent to the original L2-norm BN (L2BN) by multiplying a scaling factor which  equals to $\sqrt{\frac{\pi}{2}}$.
Experiments on various convolutional neural networks (CNNs) and generative adversarial networks (GANs) reveal that L1BN maintains almost the same accuracies and convergence rates compared to L2BN but with higher computational efficiency.
On FPGA platform, the proposed signum and absolute operations in L1BN can achieve 1.5$\times$ speedup and save 50\% power consumption, compared with the original costly square and root operations, respectively.
This hardware-friendly normalization method not only surpasses L2BN in speed, but also simplify the hardware design of ASIC accelerators with higher energy efficiency. Last but not the least, L1BN promises a fully quantized training of DNNs, which is crucial to future adaptive terminal devices.

\end{abstract}


\IEEEpeerreviewmaketitle

\section{Introduction}
Nowadays, deep neural networks (DNNs) \cite{lecun2015deep} are successfully permeating various real-life applications, for instance, computer vision \cite{krizhevsky2012imagenet}, speech recognition \cite{amodei2016deep},
machine translation \cite{wu2016google}, Go game \cite{silver2017mastering}, and multi-model tasks across them \cite{karpathy2014deep}.
However, training DNNs is complicated and needs elaborate tuning of various hyperparameters, especially on large datasets where the training samples are so variant.
The distribution shift of mini-batch inputs will affect the outputs of following layers successively and eventually make the network's outputs and gradients vanish or explode.
This internal covariate shift phenomenon leads to slower convergence in training, requires careful adaption of learning rate and increases the importance of appropriate parameters initialization \cite{he2015delving}.

To address this problem, batch normalization (BN) \cite{ioffe2015batch} has been proposed to facilitate convergence, as well as reduce the difficulty of annealing learning rate and initializing parameters.
The randomness from the mini-batch statistics serves as a regularizer during training and improves the final accuracy.
Besides, most generative adversarial networks (GANs) also rely on BN in both the generator and the discriminator \cite{radford2015unsupervised, salimans2016improved}.
BN is proved critically to help deep generators launch a normal training process, as well as prevent generators from mode collapse which is a common failure observed in GANs.
BN performs such helpfully in training DNNs that it has almost been a standard configuration along with the activation function, e.g., the rectified linear units (ReLU), coupled as BN-ReLU, not only in various deep learning models \cite{he2016deep, huang2017densely, howard2017mobilenets}, but also in the deep learning accelerator community \cite{zhao2017accelerating, jiang2017xnor, tao2017benchip}.


Inspired by BN, weight normalization \cite{salimans2016weight} uses the L2-norm of the incoming weights to normalize the summed inputs to a neuron. Layer normalization \cite{ba2016layer} transposes the statistics of a training batch to all of the summed inputs in a single training case, which do not re-parameterize the original network. Both methods eliminate the dependencies among the examples in a mini-batch and can be applied successfully to recurrent models. Batch renormalization \cite{ioffe2017batch} proposes an affine transformation to ensure that the training and inference models generate the same outputs that depend on individual examples rather than the entire mini-batch.

However, the BN layer usually causes considerable training overhead for speed and energy consumption in both the forward and backward propagations during training.
On the one hand, the additional computations are quite heavy, especially for resource-limited ASIC devices.
When it comes to online learning, i.e., deploying the training process onto terminal devices, the resource problem of BN becomes more salient and has challenged its extensive applications in various scenarios.
On the other, the square and root operations introduce strong nonlinearity that makes it difficult to employ low bit-width quantization algorithms.
Although a lot of quantization methods have been proposed to reduce the memory cost and accelerate the computation \cite{hubara2016binarized, he2016effective, deng2017gated}, the BN layer still remains in float32 precision or avoided for simplicity \cite{wu2018training}.


In this paper, we propose an L1-norm BN (L1BN), in which the L2-norm variance of each neuron's activation within one mini-batch is replaced by an L1-norm variance.
By deriving the chain rule for differentiating the L1BN layer when implementing backpropagation training, we find that the costly and strongly nonlinear square and root operations can be replaced by hardware-friendly signum and absolute operations.
We prove that L1BN is actually equivalent to the original L2BN by multiplying a scaling factor $\sqrt{\frac{\pi}{2}}$.
To verify the proposed method, various experiments have been conducted on convolutional neural networks (CNNs) over datasets including Fashion-MNIST \cite{xiao2017fashion}, SVHN \cite{netzer2011reading}, CIFAR10/100 \cite{krizhevsky2009learning} and ILSVRC12 \cite{russakovsky2015imagenet}, as well as the generative adversarial networks (GANs) over CIFAR10 and LSUN-Bedroom \cite{yu2015lsun}.
The results indicate that the L1BN is able to achieve comparable accuracy and convergence speed with the L2BN, but with higher computational efficiency.
Compared to the original complex square and root operations, on FPGA platform, L1BN achieves 1.5$\times$ speedup, and 50\% power saving, respectively.
We believe that this new normalization method promises the fully quantized low-precision dataflow for training deep models and hardware-friendly learning framework for future resource-limited terminal devices.

\section{Problem Formulation}
Stochastic gradient descent (SGD), known as incremental gradient descent, is a stochastic approximation of the gradient descent optimization and iterative method for minimizing an objective function that is written as a sum of differentiable functions:
\begin{equation}
  \Theta={\rm arg}\mathop{\rm min}\limits_{\Theta}\frac{1}{N}\sum_{i=1}^N \ell(x_i,\Theta)
\end{equation}
where $x_i$ for $i=1, 2, ... N$ is the training set containing totally $N$ samples, and $\sum_{i=1}^N \ell(x_i,\Theta)$ is the empirical risk by summarizing the risk at each sample, and $\ell(x_i,\Theta)$ is typically associated with the $i$-th observation in the dataset.
When considering a mini-batch size of $m$ updates in SGD, the true gradient of the cost function $\ell$ is approximated by $m$ samples, i.e., the gradient of each iteration in SGD can be simply estimated by processing the gradient of each sample and update with the average values
\begin{equation}
  \Theta \leftarrow \Theta - \alpha \cdot \frac{1}{m}\sum_{i=1}^m\frac{\partial \ell(x_i,\Theta)}{\partial\Theta}
\end{equation}
where $\alpha$ is the learning rate. While SGD with mini-batch is simple and effective, it requires careful tuning of the model hyperparameters, especially the learning rate used in optimizer, as well as the initial values of the model parameters.
Otherwise neural network will start with most of the activations saturated and make the training hard to converge.
Another issue is that the inputs of each layer are affected by the parameters of all preceding layers, so that small changes to the network parameters amplify as the network becomes deeper. This leads to the problem that the outputs and gradients of network will easily explode or vanish.
Thus, SGD slows down the training by requiring lower learning rates and careful parameter initialization, and makes it notoriously hard to train models with saturating nonlinearities, which affects the networks' robustness and challenges its extensive applications.

In \cite{ioffe2015batch}, the authors refer to this phenomenon
as internal covariate shift, and address this problem by normalizing layer inputs.
This method draws its strength from making normalization a part of the model architecture and performing the normalization across each training mini-batch, which is termed as ``Batch normalization" (BN).
Batch normalization allows us to use much higher learning rates and be less careful about initialization. Besides, the randomness from the batch statistics serves as a regularizer during training and in many cases eliminates the need for Dropout.

\subsection{Conventional L2BN}
Specifically, BN normalizes each scalar feature independently by making it have zero mean and unit variance.
For a layer with $d$-dimensional input $x = \{x^{(1)},...,x^{(k)},..., x^{(d)}\}$, each dimension is the input of a neural activation that is normalized by
\begin{equation}
  \hat{x}^{(k)} =\frac{x^{(k)}-{\rm{E}}[x^{(k)}]}{\sqrt{{\rm{Var}}[x^{(k)}]}}
\end{equation}
where the expectation ${\rm{E}}[x^{(k)}]$ and variance ${\rm{Var}}[x^{(k)}]$ are computed over all training samples.

Usually for a mini-batch containing $m$ samples, we use $\mu_\mathcal{B}$ and $\sigma_\mathcal{B}^{2}+\epsilon$ to estimate ${\rm{E}}[x^{(k)}]$ and ${\rm{Var}}[x^{(k)}]$, respectively, which gives that
\begin{equation}
  \label{eqn3}
 \hat{x}^{(k)}_i=\frac{x_i^{(k)}-\mu_\mathcal{B}}{\sqrt{\sigma_\mathcal{B}^{2}+\epsilon}}
\end{equation}
where $\epsilon$ is a sufficiently small positive parameter for avoiding the numerical error, the mini-batch mean $\mu_\mathcal{B}$ and variance $\sigma_\mathcal{B}$  are given by
\begin{equation}
  \mu_\mathcal{B}=\frac{1}{m}\sum_{i=1}^m{x_i}
\end{equation}
and
\begin{equation}
  \sigma_\mathcal{B}=\sqrt{\frac{1}{m}\sum_{i=1}^m{(x_i-\mu_\mathcal{B})^2}}
\end{equation}
respectively. It is easy to see that $\sigma_\mathcal{B}$ is calculated based on a L2-norm.

To guarantee that the transformation inserted into the network can represent the identify transform, a pair of parameters $\gamma^{(k)}$ and $\beta^{(k)}$ that scale and shift the normalized value are introduced
\begin{equation}
\label{gammaL2}
  y^{(k)} = \gamma^{(k)} \hat{x}^{(k)} + \beta^{(k)}
\end{equation}
where $\gamma$ and $\beta$ are parameters to be trained along with the original model parameters.

To implement backpropagation training when batch normalization is involved, the chain rule is derived as follows:
\begin{equation}
  \label{derivative1}
  \begin{array}{lll}
    \frac{\partial \ell}{\partial \hat{x}_i} = \frac{\partial \ell}{\partial y_i} \cdot \gamma  \\\\

    \frac{\partial \ell}{\partial \sigma_\mathcal{B}^{2}} = \sum_{i=1}^{m} \frac{\partial \ell}{\partial \hat{x}_i} \cdot (x_i-\mu_\mathcal{B})  \cdot \frac{-1}{2}  {(\sigma_\mathcal{B}^{2}+\epsilon)}^{-3/2}\\\\

    \frac{\partial \ell}{\partial \mu_\mathcal{B}} = \sum_{i=1}^{m} \frac{\partial \ell}{\partial \hat{x}_i} \cdot  \frac{-1}{\sqrt{\sigma_\mathcal{B}^{2}+\epsilon}} \\\\

    \frac{\partial \ell}{\partial x_i} = \frac{\partial \ell}{\partial \hat{x}_i} \cdot \frac{1}{\sqrt{\sigma^2_\mathcal{B}+\epsilon}}  +
    \frac{\partial \ell}{\partial \sigma^2_\mathcal{B}} \cdot     \frac{2(x_i-\mu_\mathcal{B})}{m} +
    \frac{\partial \ell}{\partial \mu_\mathcal{B}} \cdot \frac{1}{m} \\\\

    \frac{\partial \ell}{\partial \gamma} = \sum_{i=1}^{m}   \frac{\partial \ell}{\partial y_i} \cdot \hat{x}_i  \\\\

    \frac{\partial \ell}{\partial \beta} = \sum_{i=1}^{m} \frac{\partial \ell}{\partial y_i} \\\\
  \end{array}
\end{equation}

Although BN accelerates the convergence of training DNNs, it requires additional computation and generally slows down the training process by a large margin.
Table~\ref{tab0} shows speed tests of training with or without BN.
All models are built on Tensorflow \cite{abadi2016tensorflow} and trained on one or two Titan-Xp GPUs. Section~\ref{CNNtest} will detail the implementation and training hyperparameters.
For BN, we apply it's accelerated version: fused batch normalization.
Note that since the L2-norm is applied to estimate $\sigma_\mathcal{B}$, we term this method as the L2-norm Batch normalization (L2BN).
It is seen that, in L2BN, the BN layer requires square and root operations during the training phase.
These operations are  costly, especially on resource-limited ASIC devices. Besides, the square and root operation introduce strong nonlinearity that makes the bit-width quantization difficult. To address this issue, we propose an L1-norm batch normalization (L1BN) in this work.

\renewcommand\arraystretch{1.3}
\begin{table}[h]
\caption{Training samples per second (FPS) without or with batch normalization}
\begin{center}
\begin{tabular}{ p{2.5cm}<{\centering}
                 p{1.5cm}<{\centering}
                 p{1.5cm}<{\centering}
                 p{1.5cm}<{\centering}
                 }
\toprule
Model & no BN & with BN & time(\%)  \\
\hline
ResNet-110   & 2174 & 1480 & 31.9 \\
DenseNet-100 & 926  & 633  & 31.6 \\
MobileNet    & 645  & 512  & 20.6 \\
\bottomrule
\end{tabular}
\end{center}
\label{tab0}
\end{table}

\subsection{Proposed L1BN}
Our idea is very simple, which is just to apply the L1-norm to estimate $\sigma_\mathcal{B}$.
In Theorem 1, it will be proven that the L1BN is equivalent to L2BN by multiplying a scaling factor $\sqrt{\frac{\pi}{2}}$.
The simulation results will further show that L1BN maintains the same accuracy and convergence speed compared to L2BN but with higher computational efficiency.
The L1BN is formulated as
\begin{equation}
\label{gammaL1}
  y_i=\gamma  \cdot  \hat{x}_i+\beta
\end{equation}
and
\begin{equation}
  \hat{x}_i=\frac{x_i-\mu_\mathcal{B}}{\sigma_\mathcal{B}+\epsilon}
  \label{expressionxi}
\end{equation}
where $\gamma$, $\beta$, $\mu_\mathcal{B}$ and $\epsilon$  have the same meaning as in L2BN,
whereas $\sigma_\mathcal{B}$ is a term calculated with the L1-norm of mini-batch inputs as follows:
\begin{equation}
\label{L1BNvariance}
  \sigma_\mathcal{B}=\frac{1}{m}\sum_{i=1}^m{|x_i-\mu_\mathcal{B}|}
\end{equation}
The motivation of the replacement of the L2-norm variance by its L1-norm version is that the L1-norm variance is strongly linear correlated with the L2-norm variance.

During training we need to back propagate the gradient of loss $\ell$ through this transformation, as well as compute the gradients with respect to the parameters of the BN transform.
To implement the backward propagation when L1-norm is involved, the chain rule is derived as follows:
\begin{equation}
  \begin{array}{lll}
  \label{derivative2}
    \frac{\partial \ell}{\partial \hat{x}_i} = \frac{\partial \ell}{\partial y_i} \cdot \gamma  \\\\

    \frac{\partial \ell}{\partial \sigma_\mathcal{B}}  =\sum_{i=1}^{m} \frac{\partial \ell}{\partial \hat{x}_i} \cdot (x_i-\mu_\mathcal{B})  \cdot   \frac{-1} {(\sigma_\mathcal{B}+\epsilon)^2}\\\\

    \frac{\partial \ell}{\partial \mu_\mathcal{B}} = \sum_{i=1}^{m} \frac{\partial \ell}{\partial \hat{x}_i} \cdot   \frac{-1}{\sigma_\mathcal{B}+\epsilon} +   \frac{\partial \ell}{\partial \sigma_\mathcal{B}} \cdot \frac{-1}{m} \sum_{i=1}^m {\rm{sgn}}(x_i-\mu_\mathcal{B}) \\\\

    \frac{\partial \ell}{\partial x_i}   = \frac{\partial \ell}{\partial \sigma_\mathcal{B}} \cdot \frac{1}{m} \left\{ {\rm{sgn}}(x_i-\mu_\mathcal{B}) - \frac{1}{m}   \sum_{j=1}^m {\rm{sgn}}(x_j-\mu_\mathcal{B}) \right\} \\\\

    ~~~~~~~~+\frac{\partial \ell}{\partial \hat{ x}_i} \cdot \frac{1}{\sigma_\mathcal{B}+\epsilon} + \frac{\partial \ell}{\partial\mu_\mathcal{B}} \cdot  \frac{1}{m}

    \end{array}
  \end{equation}
and $ \frac{\partial \ell}{\partial \gamma}$, $\frac{\partial \ell}{\partial \beta}$ has exactly the same form as in Equation~\ref{derivative1}.


It is easy to see that
\begin{equation}
 {\rm{sgn}}(\hat{x}_i)={\rm{sgn}}(x_i-\mu_\mathcal{B})
\end{equation}

Let
\begin{equation}
  \begin{array}{lll}
    \mu(\frac{\partial \ell}{\partial\hat{x}_i})=\frac{1}{m}\sum_{i=1}^{m}   \frac{\partial \ell}{\partial \hat{x}_i} \\\\

    \mu(\frac{\partial \ell}{\partial\hat{x}_i} \cdot \hat{x}_i )= \frac{1}{m}\sum_{i=1}^{m} (\frac{\partial \ell}{\partial\hat{x}_i} \cdot \hat{x}_i )
  \end{array}
\end{equation}

Then, by substituting Equations~\ref{expressionxi} and \ref{L1BNvariance} into Equation~\ref{derivative2},  we can obtain that
\begin{equation}
  \begin{array}{lll}
    \frac{ \partial{\ell} }{ \partial{x_i} } = \frac{1}{\sigma_\mathcal{B}+\epsilon} \{ \frac{\partial \ell}{\partial\hat{x}_i} - \mu(\frac{\partial \ell}{\partial\hat{x}_i}) -  \\\\
    ~~~~~~~~~~~~~~~~ \mu(\frac{\partial \ell}{\partial\hat{x}_i} \cdot \hat{x}_i ) \cdot
 [{\rm{sgn}}(\hat{x}_i)-\mu({\rm{sgn}}(\hat{x}_i)] \}
  \end{array}
\end{equation}
Note that the square and root operations can be avoided for implementing forward computation and back propagation when the L1BN is involved. As seen in Experiments section, L1BN significantly reduces the computational cost compared L2BN.

\emph{Theorem 1:}
For a normally distributed random variable $X$ with mean $\mu$ and variance $\sigma$, define a random variable $Y$ such that $Y = |X-{\rm{E}}(X)|$, we have
\begin{equation}
  \frac{\sigma}{{\rm{E}}(|X-{\rm{E}}(X)|}=\sqrt{\frac{\pi}{2}}
\end{equation}

\emph{Proof:}
Note that $X-{\rm{E}}(X)$ belongs to a normal distribution with zero mean and variance $\sigma$. Then $Y=|X-{\rm{E}}(X)|$ has a folded normal distribution or half-normal distribution \cite{leone1961folded}.
Denote $\mu_Y$ as the mean of $Y$, we have
\begin{equation}
  \mu_{Y}=\sqrt{\frac{2}{\pi}} \sigma e^{\frac{-\mu^{2}}{2\sigma^2}}
\end{equation}
based on the statistical property of half-normal distribution. As $\mu=0$, we can obtain that
\begin{equation}
  \frac{\sigma}{\mu_Y}=\sqrt{\frac{\pi}{2}}
\end{equation}

\emph{Remark 1:}
By Theorem 1, let $\gamma_{L_2}$ and $\gamma_{L_1}$ be the parameters in Equation~\ref{gammaL2} and~\ref{gammaL1} for L2BN and L1BN, respectively. Ideally we have
\begin{equation}
  \gamma_{L_2}=  \sqrt{\frac{\pi}{2}} \cdot \gamma_{L_1}
\end{equation}
if the inputs of all layers belong to Gaussian distribution with zero mean. In this case, by  denoting the standard derivation  of the output of all layers for L1BN and L2BN as $\sigma_{L_1}$ and $\sigma_{L_2}$, respectively, we have
\begin{equation}
  \label{eqn_ratio}
  \sigma_{L_2}=  \sqrt{\frac{\pi}{2}}  \cdot \sigma_{L_1}
\end{equation}

The above remark is validated in the Experiments section.

\section{L1BN in MLP and CNN}
As same to L2BN, L1BN can be applied to any set of activations in networks including but not limited to multi-layer perceptrons (MLPs) and CNNs.
The only difference lies in that the L1-norm is applied to calculate   $\sigma$.
As show in Equation~\ref{eqn_ratio}, in order to compensate the difference in  standard deviation, we use the strategy as follows:

\begin{enumerate}[1)]
\item When the rescale factor $\gamma$ is not applied in batch normalization, we multiply $\sigma_{L_1}$ by $\sqrt{\frac{\pi}{2}}$ in Equation~\ref{L1BNvariance} to approximate the original $\sigma_{L_2}$.

\item When the rescale factor $\gamma$ is applied in batch normalization, we just use $\sigma_{L_1}$ and let the trainable parameters $\gamma_{L_1}$ ``learn" to compensate the difference automatically through back-propagation training.
\end{enumerate}

The other normalization processes of L1BN are the just exactly the same to that of L2BN. For an MLP, each neuron is normalized based on $m$ samples of a training batch before the activation function.
While for a CNN, different elements of the same feature map at different locations are normalized in the same way.
Similar to L2BN \cite{ioffe2015batch}, we jointly normalize all the activations in a mini-batch, over all locations.
Let $\mathcal{B}$ be the set of all values in a feature map across both the elements of a mini-batch and spatial locations, so for a mini-batch of size $m$ and feature maps of size $ h \times w$, the effective mini-batch of size is $|\mathcal{B}| = mhw $.
Then, the pair of parameters $\gamma$ and $\beta$ per feature map can be learned, rather than per activation.
So during inference the BN transform applies the same linear transformation to each neural activation in a given feature map.

The Algorithm for L1BN is presented in Algorithm~\ref{algorithm1}.

\section{Experiments}
\bigskip

\subsection{L1BN on classification tasks}
\label{CNNtest}
To evaluate the equivalence between the L1-norm batch normalization and it's original L2-norm version, we test both methods in various image classification tasks. For classification, we parameterize model complexity and task complexity, then apply two-dimensional demonstrations on multiple datasets with different CNN architectures. For each demonstration, all the  hyper-parameters between    L2BN and L1BN remain the same, the only difference is the using of L2-norm or L1-norm BN. In the following experiments, SGD with momentum 0.9 is the default optimizer setting:

\begin{algorithm}[h]
  \caption{Training a L1BN layer with statistics $\{\mu,\sigma\}$, moving-average momentum $\alpha$, trainable linear layer parameters $\{\gamma,\beta\}$, learning rate $\eta$, inference with one sample}
    \begin{algorithmic}[1]

      \REQUIRE a mini-batch of pre-activation samples for training $\mathcal{B}=\{x_{1},...,x_{m}\}$, one pre-activation sample for inference $\mathcal{I}=\{x_\text{inf}\}$
      \ENSURE updated parameters $\{ \mu, \sigma, \gamma, \beta \}$, normalized pre-activations $\mathcal{B}_\text{tr}^\text{N}=\{x_{1}^\text{N},...,x_{m}^\text{N}\}$, $\mathcal{I}_\text{inf}^\text{N}=\{x_\text{inf}^\text{N}\}$

      \setstretch{1.3}
      \textbf{1. Training with mini-batch $\mathcal{B}$:}\\

      \textbf{1.1 Forward:}\\
        \STATE $\mu_{\mathcal{B}} \leftarrow \frac{1}{m}\sum_{i=1}^m x_i$
        \hfill //mini-batch mean \\
        \STATE $\sigma_{\mathcal{B}} \leftarrow \frac{1}{m}\sum_{i=1}^m |x_i-\mu_{\mathcal{B}}|$
        \hfill //mini-batch L1 variance\\
        \STATE $\hat{x}_i \leftarrow \frac{x_i-\mu_\mathcal{B}}{\sigma_\mathcal{B}+\epsilon}$
        \hfill //normalize \\
        \STATE $x_{i}^\text{N} \leftarrow \gamma\hat{x}_i + \beta \equiv \text{L1BN}_\text{tr}(x_i)$
        \hfill //scale and shift\\

      \textbf{1.2 Backward:}\\
        \STATE $\mu \leftarrow \alpha\mu + (1-\alpha)\mu_{\mathcal{B}} $
        \hfill //update statistics\\
        \STATE $\sigma \leftarrow \alpha\sigma + (1-\alpha)\sigma_{\mathcal{B}} $
        \hfill //update statistics\\
        \STATE $\gamma \leftarrow \gamma - \eta\frac{\partial\mathcal{L}}{\partial\gamma} $
        \hfill //update parameters\\
        \STATE $\beta \leftarrow \beta - \eta\frac{\partial\mathcal{L}}{\partial\beta} $
        \hfill //update parameters\\
      \textbf{2. Inference with sample $\mathcal{I}$:}\\
        \STATE $x_\text{inf}^\text{N} \leftarrow
        \frac{\gamma}{\sigma+\epsilon} \cdot x_\text{inf} + (\beta - \frac{\gamma\mu}{\sigma+\epsilon})
        \equiv \text{L1BN}_\text{inf}(x_\text{inf}) $\\
    \end{algorithmic}
\label{algorithm1}
\end{algorithm}

1) \textbf{Simple task with shallow model:} Fashion-MNIST \cite{xiao2017fashion} is a MNIST-like fashion product database that contains 70k grayscale 28$\times$28 images and preferably represents modern CV tasks. We use a variation of LeNet-5 \cite{lecun1998gradient} with 32C5-MP2-64C5-MP2-512FC-10Softmax. The learning rate $\eta$ is set to 0.1 and divided by 10 at epoch 30 and epoch 60. The average accuracy of 10 runs on the test set is reported. As for SVHN \cite{netzer2011reading} dataset, we use a VGG-like network \cite{simonyan2014very} with totally 7 layer: 2×(128C3)-MP2-2×(256C3)-MP2-2×(512C3)-MP2-1024FC-10Softmax. The original
images are scaled and biased to the range of $[-1,+1]$, training epochs are reduced to 40 since it is a rather big dataset. The learning rate $\eta$ is set to 0.1 and divided by 10 at epoch 20 and epoch 30.

2) \textbf{Moderate tasks with very deep model:} Residual blocks \cite{he2016deep} and densely connected blocks \cite{huang2017densely} are proven to be quite efficient in very deep CNNs with much fewer weights. In order to test the effects brought by L1-norm in these structures on CIFAR \cite{krizhevsky2009learning} datasets, we train a standard 110-layer ResNet and a Wide-DenseNet (L=40, K=48) with bottle-neck layers for CIFAR10, as well as a standard 100-layer DensetNet for CIFAR100. In both datasets, images are firstly channel-wise normalized and then follow the data augmentation in \cite{lee2015deeply} for training: 4 pixels are padded on each side, and a 32$\times$32 patch is randomly cropped from the padded image or its horizontal flip. For testing, only single view of the original 32$\times$32 image is evaluated. Learning rates and annealing methods are the same as that in the original papers.

3) \textbf{Complicated task with deep wide model:} For ILSVRC12 \cite{russakovsky2015imagenet} dataset with 1000 categories, we adopt AlexNet \cite{krizhevsky2012imagenet} model but remove dropout and replace local response normalization layers with L1 or L2 batch normalization layers. Images are first rescaled such that the shorter sides are of length 256, and then cropped out centrally to 256$\times$256. For training, images are then randomly cropped to 224$\times$224 and horizontally flipped. For testing, the single center crop in validation set is evaluated. The model is trained with mini-batch size of 256 and totally 80 epochs. The weight decay is set to 5e-4, learning rate is set to 1e-2 and divided by 10 at epoch 40 and epoch 60.

\renewcommand\arraystretch{1.3}
\begin{table}[h]
\caption{Test or validation error rates for L1-norm and L2-norm batch normalization (\%)}
\begin{center}
\begin{tabular}{ p{1.5cm}<{\centering}
                 p{2.5cm}<{\centering}
                 p{1.5cm}<{\centering}
                 p{1.5cm}<{\centering}
                 }
\toprule
Dataset & Model & L2BN & L1BN   \\
\hline
Fashion  & LeNet-5       & 7.66 & \textbf{7.62} \\
SVHN     & VGG-7         & 1.93 & \textbf{1.92} \\
CIFAR10  & ResNet-110    & 6.24 & \textbf{6.12} \\
CIFAR10  & Wide-DenseNet & \textbf{4.09} & 4.13 \\
CIFAR100 & DenseNet-100  & 22.46& \textbf{22.38}\\
ImageNet & AlexNet       & 42.5 & \textbf{42.1} \\
ImageNet & MobileNet     & \textbf{29.5} & 29.7 \\
\bottomrule
\end{tabular}
\end{center}
\label{tab1}
\end{table}

4) \textbf{Complicated task with deep slim model:} Recently compact convolutions such as group convolution \cite{xie2017aggregated} and depthwise convolution \cite{chollet2016xception} draws extensive attention with fewer parameters and operations with the same performance. Therefore, we further reproduce the MobileNet \cite{howard2017mobilenets} and evaluate L1-norm batch normalization on ILSVRC12 dataset. At this time, weight decay decreases to 4e-5, the learning rate is set to 0.1 initially and linearly anneals to 1e-3 after 60 epochs. We apply the Inception data argumentation defined in TensorFlow-Slim image classification model library \cite{tfslim}. The training is performed on two Titan-Xp GPUs and the population statistics $\{\mu,\sigma\}$ for batch normalization are updated according to calculations from single GPU, so the equivalent batchsize for BN is 128.

The main results on multiple datasets with different CNN architectures are shown in Table~\ref{tab1}, all error rates are the average of best accuracies in multiple repeated experiments. Besides, the training curves of two methods in ResNet-110 are shown in Figure~\ref{fig2}. We have two major observations:

(1) From the perspective of the final results, L1-norm BN is equal to or slightly surpasses the original L2-norm BN in many occasions.
We argue that this performance improving is caused by the enlarged variation statistics: according to Equation~\ref{eqn_ratio}, the ``standard derivation'' of L1-norm is 1.25$\times$ greater than that of L2-norm.
Although the following linear layer of BN can ``learn'' to alleviate the scaling by adjusting parameter $\gamma$, L1-norm somehow brings in an additional form of randomness and may further regularize large model and finally perform better performance.
Section~\ref{layerwise} will detail the empirical explanation and give more evidences.
As for MobileNet, this compact model has less trouble with overfitting. Besides, there are so few parameters in its depthwise filters. The result in Table~\ref{tab1} indicates that the additional L1-norm regularization may hurt the accuracy by a small margin.

\begin{figure}[h]
\begin{center}
\includegraphics[width=3.5in]{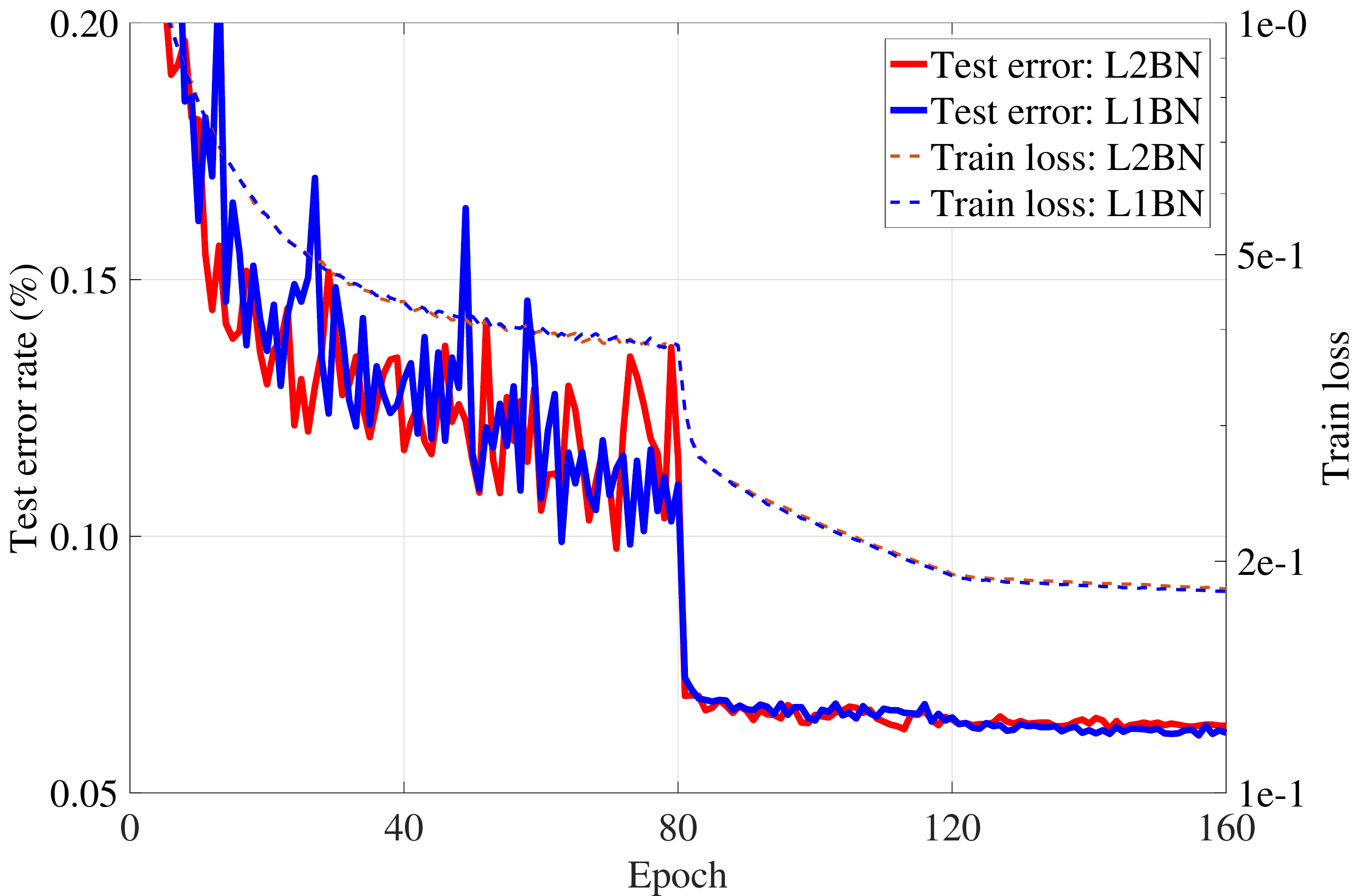}
\end{center}
\caption{Training curves of L2BN and L1BN in ResNet-110. Train losses contain weight decays of each parameters.}
\label{fig1}
\end{figure}

(2) From the perspective of training curves, the test error of L1BN is a bit more unstable at the beginning. Although this can be improved by more accurate initialization: initialize $\gamma$ from $1.0$ (L2-norm) to $0.8$ (L1-norm), the two loss curves almost overlap completely. So we directly embrace this instability and let networks find their way out. We observe no other regular distinctions of two optimization processes in all our CNN experiments.

\begin{figure*}[t]
\begin{center}
\includegraphics[width=7.1in]{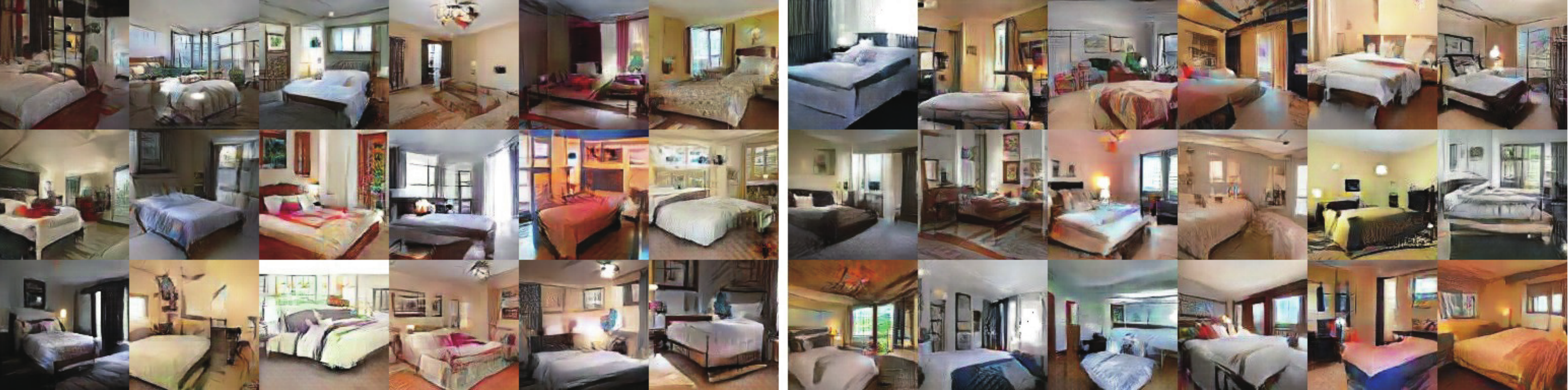}
\end{center}
\caption{Samples of 128$\times$128 LSUN bedrooms generated by a DCGAN generator using L2-norm BN (left) or L1-norm BN (right), the hyperparameters and training techniques are the same as described in WGAN-GP.}
\label{fig2}
\end{figure*}

\subsection{L1BN on generative tasks}
Since the training of GANs is a zero-sum game between two neural networks with no guarantee of convergence, most GAN implementations rely on BN in both the generator and the discriminator to help stabilize training, and prevent the generator from collapsing all samples to a single point which is a common failure mode observed in GANs. In this case, the quality of generated results will be more sensitive to any numerical or hyperparameters differences. Therefore we apply L1-norm BN to two GANs tests to prove its effectiveness.

Firstly, we generate 32$\times$32 CIFAR10 images using the same architectures as described in DCGAN \cite{radford2015unsupervised}.
For the results of DCGAN, we use the non-saturating loss function proposed in the original GAN formulation \cite{goodfellow2014generative}.
For results of WGAN-GP, the network remains the same except that no BN is applied to discriminator as suggested in \cite{gulrajani2017improved}.
Besides, the training techniques, e.g., using wasserstein distance, gradient penalty, are adopted.
Generated images are measured by the widely used metric Inception score (IS) introduced in \cite{salimans2016improved}.
Since IS fluctuates in multiple tests after every 10k generator iteration, we report both the average score of the last 20 evaluation (AVG) and the best score (BEST) of the overall training.
Note that the training techniques introduced in \cite{salimans2016improved,huang2016stacked} are not used in our implementations.
Besides, the hyperparameters and network structures are not the same.
Incorporating these techniques might further bridge the performance gap of our models.
We just used pure tests without any further fine tuning which are proven to really make a difference \cite{lucic2017gans}.
Table~\ref{tab2} shows that L1BN still equivalent to L2BN in such hyperparameter-sensetive adversarial occasions.

\renewcommand\arraystretch{1.3}
\begin{table}[h]
\caption{Unsupervised Inception scores on CIFAR-10 (larger values represent for better generation quality).}
\begin{center}
\begin{tabular}{ p{3.0cm}<{\centering}
                 p{0.5cm}<{\centering}
                 p{1.8cm}<{\centering}
                 p{1.8cm}<{\centering}
                 }
\toprule
Method & BN & AVG & BEST   \\
\hline
DCGAN (in \cite{huang2016stacked})
& L2 &\multicolumn{2}{c}{6.16 $\pm$ 0.07}\\
Improved GAN \cite{salimans2016improved}
& L2 &\multicolumn{2}{c}{6.86 $\pm$ 0.06}\\
WGAN-GP \cite{gulrajani2017improved}
& L2 &\multicolumn{2}{c}{7.86 $\pm$ 0.07}\\
\hline
\multirow{2}{*}{DCGAN (ours)}   & L2 & 7.07 $\pm$ 0.08 & 7.39 $\pm$ 0.08 \\
                                & L1 & \textbf{7.18 $\pm$ 0.09}
                                     & \textbf{7.70 $\pm$ 0.08} \\
\multirow{2}{*}{WGAN-GP (ours)} & L2 & \textbf{6.89 $\pm$ 0.12}
                                     & \textbf{7.02 $\pm$ 0.14} \\
                                & L1 & 6.86 $\pm$ 0.11 & 6.97 $\pm$ 0.11 \\

\bottomrule
\end{tabular}
\end{center}
\label{tab2}
\end{table}

Secondly, we perform experiments on LSUN-Bedrooms \cite{yu2015lsun} 128$\times$128 image generation task with all the L2-norm BN replaced with L1-norm BN.
In this case, an additional upsample de-convolution and downsample convolution is applied to DCGAN's generator and discriminator, respectively.
In order to stabilize training and avoid mode collapses, we follow the WGAN-GP method and no BN is applied to discriminator.
Since in field of GANs there is no convincing evaluating indicator or loss function to compare the generator's performance for LSUN, Figure~\ref{fig2} just intuitively shows samples of generated images after 300k generator iterations. Still, both methods generate comparable samples with detailed texture, we observe no significant differences in artistic style.

\subsection{Layerwise and channelwise comparasion}
\label{layerwise}
We further offer a layerwise and channelwise perspective to show the approximated equivalence between L1-norm and L2-norm scaling. After training a ResNet-110 for 100 epochs, we fix all the parameters and moving averaged statistics and feed the model with a batch of test images. Since the numerical difference between L1BN and L2BN will accumulate among layers, we guarantee that the inputs of each layer for both algorithms are the same. But within each individual layer, the standard deviation (L2-norm) $\sigma_{L_2}$ and the  L1-norm deviation $\sigma_{L_1}$ of channel outputs are calculated simultaneously.

\begin{figure}[h]
\begin{center}
\includegraphics[width=3.5in]{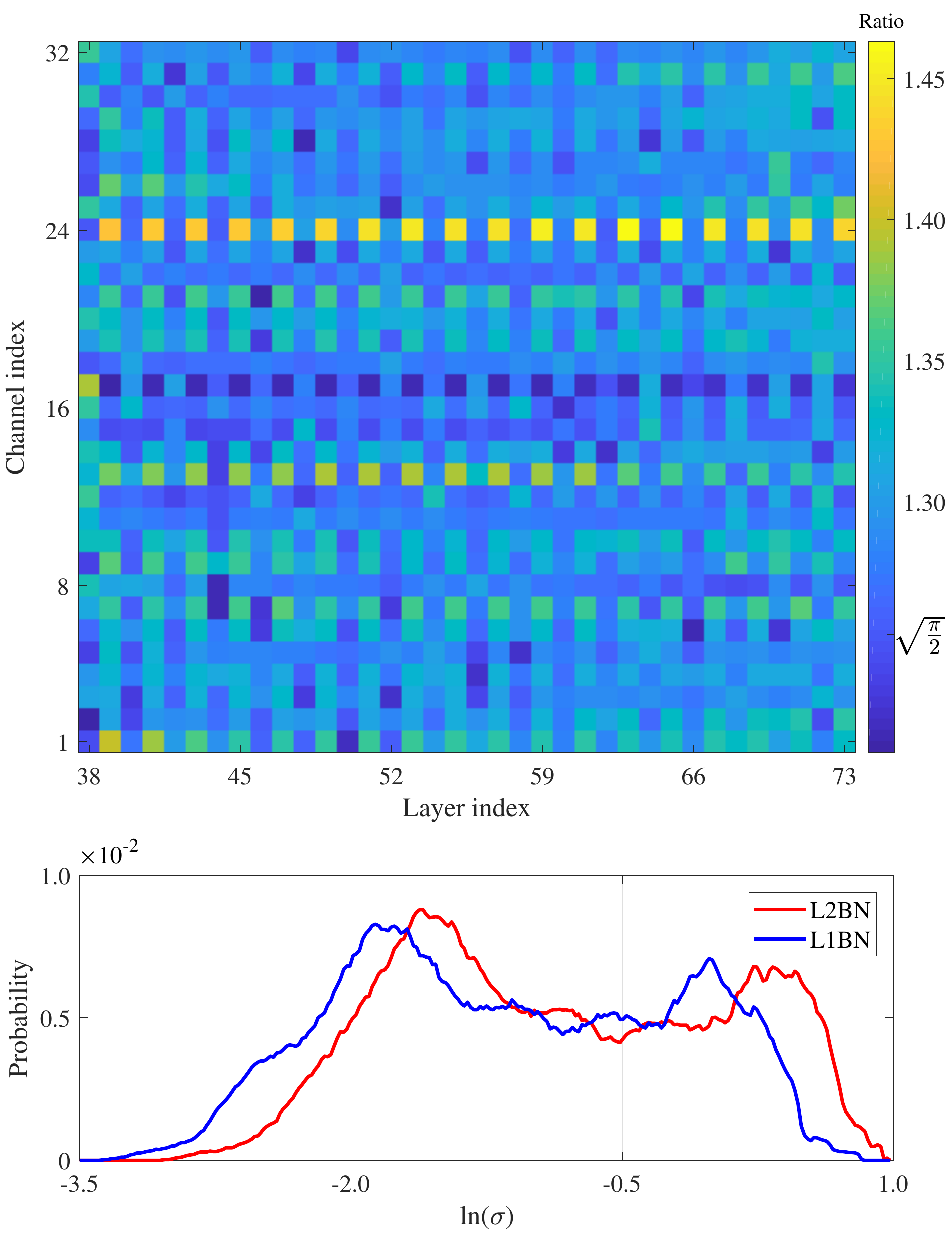}
\end{center}
\caption{Top: colormap of layerwise and channelwise ratios $\sigma_{L_2}/\sigma_{L_1}$. Ratios are averaged across 100 batches and ideally close to $\sqrt{\frac{\pi}{2}}\approx 1.25$. Down: probability histograms for $\sigma_{L_2}$ and $\sigma_{L_1}$, the axis $x$ is log-scaled.}
\label{fig3}
\end{figure}

In Remark 1, it is pointed out that, when the input signals belong to Gaussian distribution, ideally L1BN and L2BN are equivalent if we multiply  $\sqrt{\frac{\pi}{2}}$ in Equation~\ref{L1BNvariance} for L1BN. This implies that, if all the other conditions are the same, the standard derivation of the output of L2BN at each layer is $\sqrt{\frac{\pi}{2}}$ multiple of that of L1BN.
The average ratios $\sigma_{L_2}/\sigma_{L_1}$ are shown in Figure~\ref{fig3}, which demonstrates this phenomenon well. As seen the colormap in Figure~\ref{fig3}, compared to L2-norm, the L1-norm statics are theoretically smaller and will cause larger outputs. When the distribution of the data points in the intermediate layers (from layer 38 to layer 73) is relatively closer to Gaussian distribution, the ratio $\sigma_{L_2}/\sigma_{L_1}$ is close to the value of $\sqrt{\frac{\pi}{2}}\approx 1.25$. The histograms of $\sigma_{L_2}$ and $\sigma_{L_1}$ are similar except for a phase shifting in the log scale $x$ axis, which is consistent with Theorem 1 and Remark 1.
All in all,
the use of L1-norm brings in an additional form of randomness and may further regularize large model and can perform better performance.

\subsection{Computational efficiency of L1BN }
Via replacing the L2-norm variance by L1-norm variance, L1BN improves the computational efficiency. We firstly estimate the computational overhead of basic arithmetic operations on FPGA platform, all in 32-bit floating-point representation.
Table~\ref{tab3} shows the comparison of these four operations.
An Intel FPGA board (DE5-Net) is used to measure the performance and to estimate the cost of the operations based on Altera OpenCL toolchain.
Altera's PowerPlay Early Power Estimator is used to estimate power consumption.
The time and power data are reduced by the reference value of null operation.
Compared to square and root operations, signum and absolute operations saves registers and doesn't require DSP blocks.

\renewcommand\arraystretch{1.3}
\begin{table}[h]
\caption{Computational overhead of several basic arithmetic operations on FPGA}
\begin{center}
\begin{tabular}{ p{2.0cm}<{\centering}
                 p{1.1cm}<{\centering}
                 p{1.1cm}<{\centering}
                 p{1.1cm}<{\centering}
                 p{1.1cm}<{\centering}
                 }
\toprule
  &sign & abs & square  & root  \\
\hline
Registers    &153 &337  &407 &438\\
DSP blocks   &0 &0 &1 &2 \\
time (ns) &1  &1 &3 &28 \\
power (\textmu W) &2 &6 &15 &40 \\
\bottomrule
\end{tabular}
\end{center}
\label{tab3}
\end{table}

\begin{figure}[t]
\begin{center}
\includegraphics[width=3.5in]{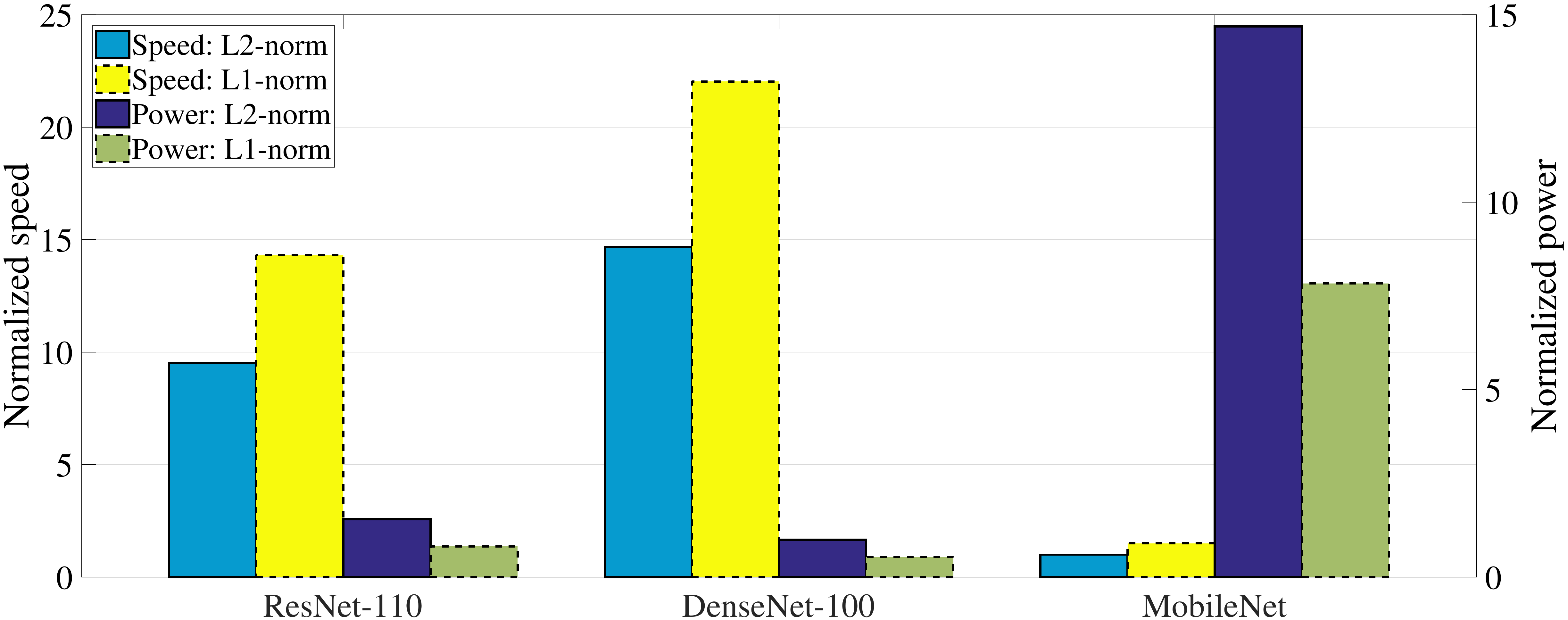}
\end{center}
\caption{Estimated time and power consumption for L1-norm and L2-norm on different CNN models.}
\label{fig4}
\end{figure}

Secondly, the time and power consumption of L1-norm and L2-norm on state-of-art CNN models are estimated in Figure~\ref{fig4}.
We count the total number of signum, absolute, square and root operation according to Equation~\ref{eqn3},~\ref{derivative1},~\ref{L1BNvariance},~\ref{derivative2}. Then weight them with power and time statistics in Table~\ref{tab3}.
Since the root operation is performed only once for each channel, we can omit its power and time consumption and aggressively compare the sum of signum and absolute operations with the square operations in most MLPs and CNNs.
The L1-norm can approximately achieve 1.5$\times$ speedup and 50\% power saving in practice.
We can conclude that by removing the complex square and root operations in conventional L2BN layer, L1BN is able to greatly improve the training efficiency on overhead of hardware resources, time and power consumption, especially for resource-limited mobile devices where DSP or FPU are not available.




\section{Discussions and Conclusions}
To minimize the cost of conventional L2-norm based batch normalization layer, we propose the L1-norm based batch normalization.
By removing the costly square and root operations during training, L1BN enables higher computational efficiency.
We derive the chain rule for differentiating the L1BN layer when implementing backpropagation training, and it is shown that the costly and strongly nonlinear square and root operations can be replaced by hardware-friendly absolute and signum operations.
We also prove that L1BN is equivalent to L2BN by multiplying a scaling factor of $\sqrt{\frac{\pi}{2}}$ when the inputs obey a Gaussian distribution with zero mean, which is commonly designed in conventional deep neural works.
Various structures of CNNs and GANs, and different datasets are tested, which present comparable classification accuracy, generation quality and convergence rate.
Cost comparisons of basic operations are estimated on FPGA platform, the L1-norm operations are able to obtain 1.5$\times$ and 50\% energy saving as well.
Other hardware resources, such as register and DSP blocks, can also be reduced.

Previous accelerators mainly target at the off-line inference, i.e., the deployment of a well-trained compressed network.
Wherein the multiply-accumulate (MAC) operations usually occupy most attention whereas the costly BN can be regarded as a scale and bias layer once training is done.
However, the capability of continual learning in real-life and on-site occasions is essential for future artificial general intelligence (AGI).
Where an autonomous agent has to explore and exploit, accumulate experiences, summarize and extract knowledge, and eventually evolve higher-level intelligence.
So online training is very import for both the data center over the cloud equipped with thousands of CPUs, GPUs as well as on edge devices with dedicated resource-limited FPGA, ASIC, wherein the BN should not be bypassed.
The advantages of L1BN with less resource overhead, faster speed, and lower energy, not only benefit most of the current deep learning models, but also make the highest ideal (AGI) one step closer.

On the other side, transferring both training and inference processes to low-precision representation is an effective leverage to alleviate the burden of hardware design.
Regretfully, most existing quantizing methods also remain the BN layer in full-precision (float32) in both the forward and backward passes because of the strongly nonlinearity of square and root operations.
Via replacing these complex operations by absolute and signum operations, L1BN greatly promises the fully quantized neural networks with low-precision dataflow for efficient online training, which is crucial to future adaptive terminal devices.


%
%

\balance

\bibliography{L1BN}
\bibliographystyle{IEEEtran}

\end{document}